\documentclass[runningheads]{llncs}

\usepackage{eccv}

\usepackage{eccvabbrv}

\usepackage{graphicx}
\usepackage{booktabs}
\usepackage{color, colortbl}
\definecolor{Gray}{gray}{0.9}
\usepackage{mathtools}
\usepackage{amsmath,amssymb}
\usepackage{multirow}
\usepackage{xpatch}
\usepackage{tabularx}
\usepackage[export]{adjustbox}
\usepackage{adjustbox}
\usepackage{array}
\usepackage{subcaption, lipsum}

\newcolumntype{R}[2]{%
    >{\adjustbox{angle=#1,lap=\width-(#2)}\bgroup}%
    l%
    <{\egroup}%
}

\usepackage[accsupp]{axessibility}  

\usepackage{hyperref}

\usepackage{orcidlink}

\begin{document}

\title{DIAL: Dense Image-text ALignment \\for Weakly Supervised Semantic Segmentation} 

\titlerunning{Dense Image-text ALignment for Weakly Supervised Semantic Segmentation}

\author{Soojin Jang\inst{1}\orcidlink{0000-0002-2719-7646} \and
Jungmin Yun\inst{2}\orcidlink{0000-0001-6868-286X} \and
Junehyoung Kwon\inst{2}\orcidlink{0000-0001-7887-5884} \and 
Eunju Lee\inst{1}\orcidlink{0000-0002-6571-0156} \and 
Youngbin Kim\inst{1,2}\orcidlink{0000-0002-2114-0120}}

\authorrunning{S.~Jang et al.}

\institute{Graduate School of Advanced Imaging Science, Multimedia \& Film,\\ Chung-Ang University, Korea \and
Department of Artificial Intelligence, Chung-Ang University, Korea
\email{\{sujin0110,cocoro357,dirchdmltnv,dmswn5829,ybkim85\}@cau.ac.kr}}

\maketitle

\begin{abstract}
Weakly supervised semantic segmentation (WSSS) approaches typically rely on class activation maps (CAMs) for initial seed generation, which often fail to capture global context due to limited supervision from image-level labels. To address this issue, we introduce DALNet, Dense Alignment Learning Network that leverages text embeddings to enhance the comprehensive understanding and precise localization of objects across different levels of granularity. Our key insight is to employ a dual-level alignment strategy: (1) Global Implicit Alignment (GIA) to capture global semantics by maximizing the similarity between the class token and the corresponding text embeddings while minimizing the similarity with background embeddings, and (2) Local Explicit Alignment (LEA) to improve object localization by utilizing spatial information from patch tokens. Moreover, we propose a cross-contrastive learning approach that aligns foreground features between image and text modalities while separating them from the background, encouraging activation in missing regions and suppressing distractions. Through extensive experiments on the PASCAL VOC and MS COCO datasets, we demonstrate that DALNet significantly outperforms state-of-the-art WSSS methods. Our approach, in particular, allows for more efficient end-to-end process as a single-stage method.
\end{abstract}

\section{Introduction}
\label{sec:intro}
Semantic segmentation is the task of assigning a semantic label to each pixel in an image. While training supervised models for such tasks demands labor-intensive pixel-level annotations, weakly supervised semantic segmentation (WSSS) addresses this challenge by learning to segment objects using class labels assigned to the entire image~\cite{ru2023token, wang2020self, hariharan2011semantic,du2022weakly,zhang2020reliability}.

\begin{figure}[t!]
  \centering
    \includegraphics[width=1.0\linewidth]{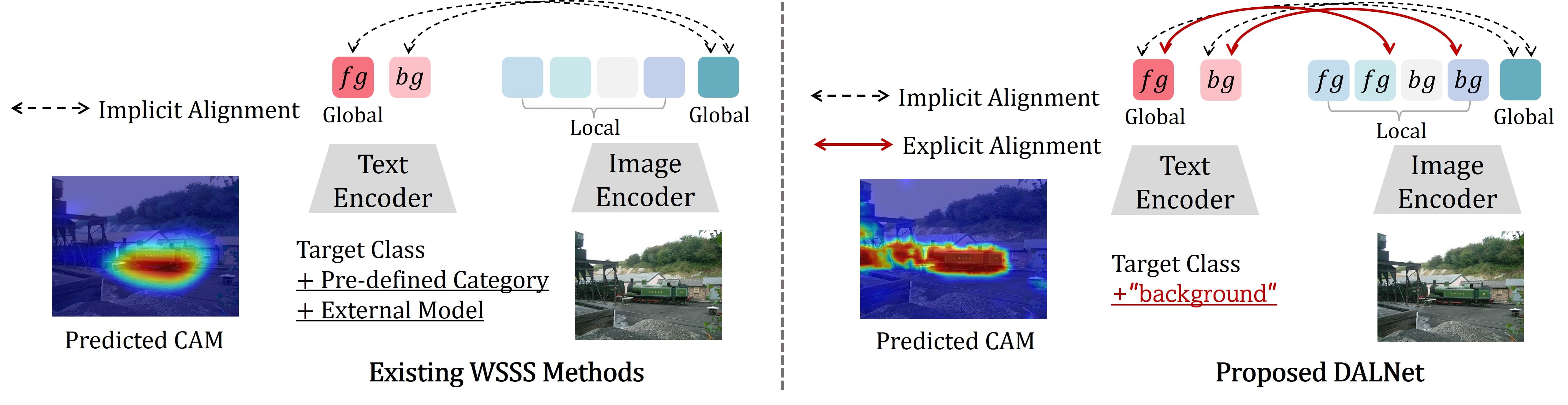}
   \caption{Comparison of \textbf{(Left)} existing WSSS methods and \textbf{(Right)} proposed DALNet. \textbf{(Left)} Existing methods for implicit alignment depend on global image features, potentially missing local details within the image. \textbf{(Right)} In contrast, the proposed DALNet integrates global and local features to preserve spatial details and facilitate explicit alignment. It distinguishes between foreground and background in image patches and text, addressing various levels of granularity. The dual alignment mechanism captures diverse object regions without any pre-defined category or external model.}
   \label{fig:fig1}
\end{figure}

In most WSSS methods with image-level class labels, class activation maps (CAMs)~\cite{zhou2016learning} are used as the initial seeds, refined to generate pseudo labels, and finally trained segmentation with the pseudo labels. However, CAMs often struggle to accurately identify object regions, primarily relying on convolutional neural networks (CNNs) with inherent limitations in contextual understanding, hindering the activation of object regions in images~\cite{lee2021railroad, li2022expansion, jiang2022l2g}. Several studies have proposed improving CAMs using vision transformer (ViT) architecture~\cite{dosovitskiy2020image}, recognized for capturing global relationships, offering a potential improvement over CAMs in addressing WSSS challenges~\cite{ru2023token, xu2022multi, ru2022learning}. 

Despite these advancements, relying solely on image-level class labels for WSSS can lead to incomplete object activation in CAMs. To address this issue, several language-guided models~\cite{radford2021learning, xu2023learning} have emerged, leveraging natural language to emphasize visual relationships and object categories. Recent studies on WSSS~\cite{xie2022clims,lin2023clip,xu2023learning} have incorporated text supervision to address the challenges posed by limited guidance. These approaches have adopted contrastive language-image pre-training (CLIP)~\cite{radford2021learning} to extract text representations from linguistic descriptions. CLIP employs a contrastive learning strategy to align images with their corresponding texts, distinguishing them from other texts. However, existing WSSS methods often depend on implicit alignment, aggregating representations such as global image features, and neglecting the spatial details, as illustrated in \cref{fig:fig1} (Left)~\cite{xie2022clims,lin2023clip}. These methods only implicitly align visual and textual concepts at the same level of granularity~\cite{chen2023revisiting}, limiting the localization performance~\cite{yang2022vision}. In addition, these methods often rely on pre-defined sets of background categories or external models to extract background text representations.

To overcome such limitations, we propose \textbf{Dense Alignment Learning Network (DALNet)}, which employs a dual-level alignment mechanism for vision-language representation learning in WSSS. As shown in ~\cref{fig:fig1} (Right), we consider various levels of granularity by differentiating between global and local attributes using class and patch tokens, further dividing them into foreground and background for both image and text. Firstly, we implement \textbf{Global Implicit Alignment (GIA)} to capture the rich semantics of the image from a global perspective. This approach measures the similarity between the class token, which aggregates the global features of the image, and text embeddings that distinguish between the foreground target class and the background. Secondly, we propose \textbf{Local Explicit Alignment (LEA)} to enhance object localization by utilizing spatial details within patch tokens. We identify visual features for foreground and background patch tokens using an object-aware mask and then evaluate their similarity to corresponding text embeddings. This method facilitates explicit alignment in the vision-language embedding space through image-text cross-contrastive learning. Unlike existing methods, our novel cross-contrastive learning approach uniquely considers positive and negative pairs across both foreground and background representations. It adjusts the foreground in images to correspond with target classes, thereby mitigating over-activation in irrelevant areas and preventing the over-activation of excessively specific areas. Leveraging this granular information improves comprehension and object localization, while also activating missing regions and reducing distractions. Our method efficiently learns foreground and background representations using only class information and the ``\texttt{background}'' term, providing practical advantages without relying on additional prior knowledge.

Furthermore, unlike most existing WSSS approaches that involve multi-stage processes~\cite{xie2022clims,jang2023weakly,jiang2022l2g,zhou2022regional}, we integrate DALNet into a single-stage WSSS that shares the encoder for classification and segmentation networks, enabling end-to-end training. The main contributions of this paper are summarized as follows:

\begin{itemize}
  \item We introduce Dense Alignment Learning Network (DALNet) for WSSS, which integrates global and local visual features through a dual-level alignment strategy. Global Implicit Alignment (GIA) and Local Explicit Alignment (LEA) are designed to enhance dense localization and comprehensive understanding of various objects within the image.
  \item We propose a novel cross-contrastive learning approach that aligns visual features with text embeddings by employing contrastive learning across an image's foreground and background at various levels of granularity. Using only the target class and the ``\texttt{background}'' term, this approach effectively activates relevant regions while suppressing irrelevant objects.
  \item Experimental results on the PASCAL VOC and MS COCO datasets demonstrate the effectiveness of our proposed method, surpassing state-of-the-art single-stage WSSS methods and exhibiting competitive performance comparable to multi-stage methods.
\end{itemize}

\section{Related Work}
\subsection{Weakly Supervised Semantic Segmentation}
Several WSSS methods have employed CNNs to generate CAMs as initial seeds~\cite{ahn2018learning, ahn2019weakly, kweon2023weakly, chen2022self, lee2021anti, wang2020self, jang2023weakly}. However, recent research suggests that CNN-based architectures are limited in capturing global information effectively due to their restricted receptive fields~\cite{veit2016residual, kim2023dead, ding2022scaling}. Various approaches have been proposed to address the locality problem of the CNN architecture~\cite{wei2018revisiting, kim2021discriminative, kim2021discriminative, li2022expansion, jiang2022l2g}. Recently, WSSS methods have adopted ViT as a backbone for generating localization maps~\cite{xu2022multi, ru2023token, ru2022learning}. By leveraging the multi-head self-attention mechanism, ViT effectively models long-range dependencies by gathering information from diverse regions, surpassing CNNs~\cite{raghu2021vision, vaswani2017attention}. MCTFormer uses a data-efficient variant of ViT, DeiT-S~\cite{touvron2021training}, as its backbone to model interactions between multiple class tokens and patch tokens~\cite{xu2022multi}. Moreover, L2G makes use of both global and local contexts to obtain more integral object attention~\cite{jiang2022l2g}.

To improve efficiency, various approaches have focused on single-stage WSSS. AFA learns reliable semantic affinity and refines the initial pseudo labels with low-level image appearance~\cite{ru2022learning}. In addition, ToCo addresses over-smoothing and enhances semantic consistency by contrasting class tokens between the local and global regions of ViT~\cite{ru2023token}. Moreover, ViT-PCM preserves the locality characteristics of ViT and performs effective mapping between multi-label classification and semantic segmentation~\cite{rossetti2022max}. 

\subsection{Vision-language Pre-training}
Recently, visual-language pre-training (VLP) models addressing large-scale image-text pairs have demonstrated robust performance in downstream tasks~\cite{desai2021virtex, jia2021scaling, li2019visualbert, lu2019vilbert, radford2021learning, alayrac2022flamingo}. Additionally, several approaches have focused on learning image representations with language supervision for dense prediction tasks~\cite{xu2022groupvit, yun2023ifseg, yi2023simple}. Open-vocabulary semantic segmentation has been explored, which segments images based on arbitrary categories described through texts~\cite{mukhoti2023open, xu2022groupvit, wang2022cris}. However, these methods often depend on fine-tuning modules or require segmentation annotations and extensive external training datasets. Some recent studies have integrated visual-language models and language supervision into WSSS~\cite{xie2022clims, xu2023learning,lin2023clip}. CLIMS employs the visual-language model~\cite{radford2021learning} and language-guided supervision to effectively identify object regions while suppressing background regions~\cite{xie2022clims}. Additionally, class representation capabilities have been enhanced through the exploitation of language priors and class-specific tokens~\cite{xu2023learning}. CLIP-ES employs text-driven strategies by modifying text input to capitalize on the strengths of CLIP~\cite{lin2023clip}. Similarly, we introduce DALNet based on language-guided supervision. In contrast to other approaches, the proposed method can generate text embeddings without requiring additional models or pre-defining the related text. Furthermore, we employ GIA and LEA to address the limitations of aggregated features in existing methods, leading to poor object localization.

\subsection{Contrastive Learning for Segmentation Tasks}
Contrastive learning aims to train the model to bring similar samples close together and dissimilar samples far apart in the given input data without explicit labels for training~\cite{chen2020simclr, he2020momentum, grill2020bootstrap}. This approach involves using pairs of identical samples as positives and pairs of different samples as negatives to learn the similarity between the data. Recently, several researches have aimed to enhance localization performance by leveraging the alignment between visual and text embeddings during segmentation~\cite{xu2022groupvit, liang2023open}. GroupViT groups similar visual concepts by applying contrastive loss between pooled image tokens and text representations~\cite{xu2022groupvit}. In addition, research has been conducted on open-vocabulary segmentation by contrasting masked image regions with nouns using CLIP~\cite{liang2023open}. Contrastive learning has been introduced into WSSS to bridge the information gap between weak labels and target labels~\cite{du2022weakly, xie2022c2am, zhou2022regional, kwon2024learning}. An approach has been introduced for contrasting pixels and prototypes to improve the performance of the localization map~\cite{du2022weakly}. C2AM disentangles foreground and background attributes within an image by leveraging contrastive learning~\cite{ xie2022c2am}. RCA employs contrastive loss to increase similarity in the same class for region-aware representation while decreasing similarity for different classes~\cite{zhou2022regional}. SMA leverages contrastive learning to separate distinct features in order to synthesize diverse combinations of object-background representations~\cite{kwon2024learning}. Unlike existing methods, the proposed method employs cross-contrastive learning between images and texts, considering positive and negative pairs across various granularities. This approach our proposed DALNet to activate in missing regions while suppressing distractions and focusing on the target objects.

\section{Methodology}
\subsection{Method Overview}
\begin{figure*}[!t]
      \centerline{\includegraphics[width=12cm]{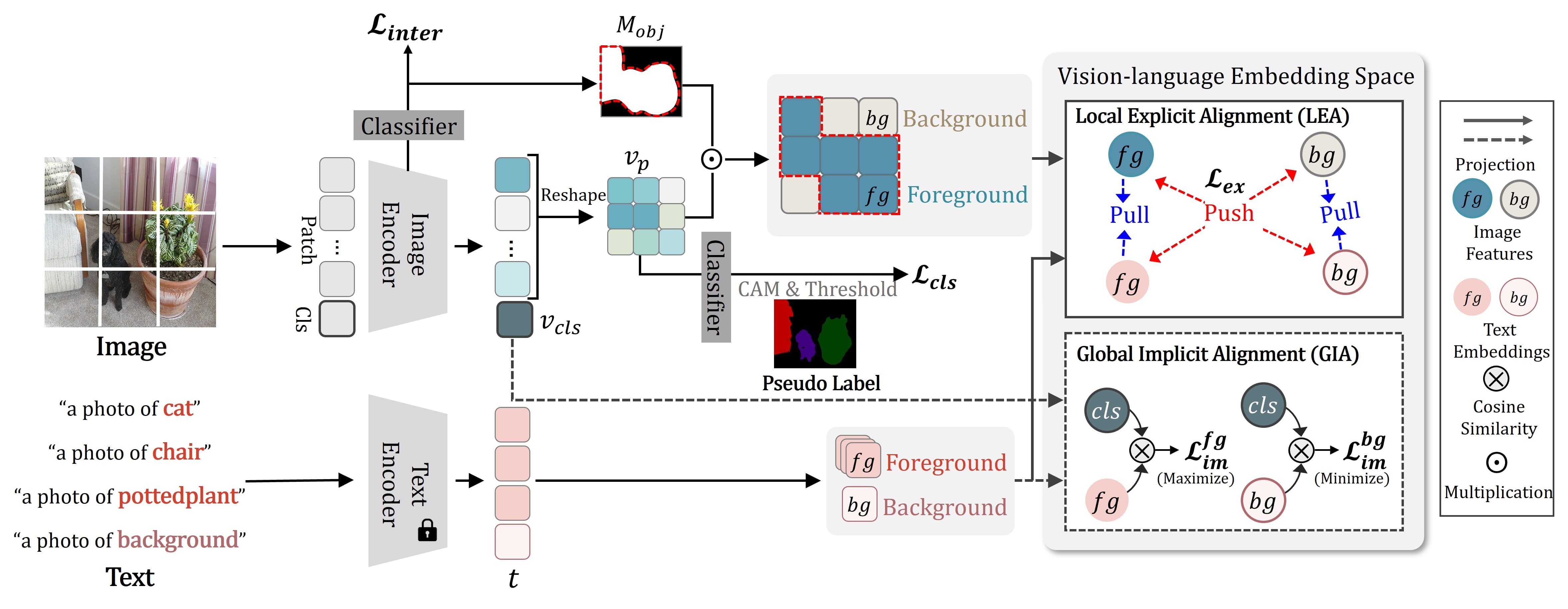}}

    \caption{\textbf{Overview of the proposed DALNet.} This approach employs a visual encoder to extract features and an object-aware mask $M_{obj}$ to distinguish foreground and background. Text prompts are fed into a text encoder to generate embeddings for target objects and the background. Cross-contrastive learning aligns representations from both modalities, associating each token with either the foreground or background. GIA contrasts the class token with text embeddings to incorporate global information, while LEA leverages patch tokens and text embeddings for precise localization.}
    \label{fig:fig2} 
\end{figure*}

The overview of the proposed method is illustrated in \cref{fig:fig2}. Visual and textual features are extracted from the image encoder and text encoder, respectively. Notably, we employ simple prompt templates, ``a photo of \texttt{[CLS]}'' and ``a photo of \texttt{background},'' to obtain text embeddings associated with foreground and background, respectively. We utilize intermediate features from the image encoder to generate an object-aware mask, which offers a rough localization of the object and background regions. Then, multiplying the object-aware mask by reshaped patch tokens generates foreground and background representations rich in regional information. We leverage cross-contrastive learning between the representations of the two modalities and employ LEA, preserving local information. Additionally, we perform GIA by maximizing the similarity between the class token obtained from the image encoder and the foreground text embeddings while minimizing the similarity with the background embeddings. Finally, a classifier produces the final CAMs to generate the final pseudo labels.

\subsection{Revisiting Class Activation Maps}
The primary objective of CAMs is to identify specific regions within images activated during the prediction process in the classification network. Due to their simplicity and efficiency, CAMs have gained extensive application as a method of generating initial seeds in WSSS. Given an image, feature maps $F \in \mathbb{R}^{HW \times D}$ are extracted from a classification network, where $HW$ denotes the size of the spatial resolution and $D$ represents channel dimensions. Within the classification network, classifier weights $W \in \mathbb{R}^{C\times D}$ are used to weight and sum the feature maps $F$, where $C$ denotes the number of classes. Negative activations can be mitigated by scaling CAMs to the range of $[0, 1]$ using the max normalization and the $\mathtt{ReLU}$ function. Thus, CAMs for $c$-th class are determined as follows:
\begin{equation}
\mathtt{CAM}_c(F,W)=\frac{\mathtt{ReLU}(A_c)}{\mathrm{max}(\mathtt{ReLU}(A_c))}, \quad
A_c = \sum_{i}W_{c,i}F_{:,i}.
\label{eq:eq1}
\end{equation}

In general, a background threshold $\beta$ is used to differentiate between the foreground and background regions within the image. 

\subsection{Image and Text Representations}
\label{sec:features}
{\bf Visual Features.} We employ ViT as a visual encoder, generating a class token that aggregates global semantics for predicting the output class, along with patch tokens that encode rich information about their corresponding local image patches~\cite{jiang2021all}. 
To generate patch tokens $x_{p} \in \mathbb{R}^{N^2 \times d_v}$, we initially split the input image into $N \times N$ patches, which are then embedded into a sequence of $N^2$ patch tokens. Here, $d_v$ represents the channel dimension. Subsequently, we use a learnable class token $x_{cls} \in \mathbb{R}^{1\times d_v}$ as aggregated representations of the entire image. We prepend the class token $x_{cls}$ to the sequence of patch tokens $x_{p}$, apply positional embedding, and then use them as input for the visual encoder. The resulting visual features $V= \left\{ {v_{p}, v_{cls}} \right\}$ from the visual encoder comprise the patch and class tokens, denoted as $v_{p} \in \mathbb{R}^{N^2\times d_v}$ and $v_{cls} \in \mathbb{R}^{1 \times d_v}$, respectively.

Furthermore, we leverage information from the intermediate layers to aggregate foreground and background attributes from visual features. To preserve semantic diversity, we leverage intermediate features that encode general class attributes rather than specific object details. Using these intermediate features, we construct an object-aware mask. Visual features $F_{inter}$ are obtained through intermediate block from the visual encoder. A global max pooling operation is applied to aggregate intermediate patch tokens, following ~\cite{ru2022learning}. Following~\cref{eq:eq1}, we compute the intermediate CAMs, employing the visual feature $F_{inter}$ and convolutional layer parameters $\theta_{inter}$. The object-aware mask $M_{obj}\in\{0,1\}^{N\times N}$ is obtained by applying the background threshold $\beta$ to intermediate CAMs. $M_{obj}$ is a binary map where the foreground region has a value of 1, and the background region has a value of 0, representing a coarse localization map of the input image. Then, we multiply the reshaped patch tokens $v_{p}\in\mathbb{R}^{N\times N\times d_v}$ by the object-aware mask $M_{obj}$ as follows:
\begin{equation}
    v_{p}^{fg} = v_{p} \odot M_{obj}, \quad
    v_{p}^{bg} = v_{p} \odot (1-M_{obj}),
\end{equation}
where $\odot$ denotes element-wise multiplication. In addition, $v_{p}^{fg}$ and $v_{p}^{bg}$ contain information about the foreground and background features, respectively. These features are used for dense alignment between image and text representations. \\
{\bf Text Embeddings.} We employ the text encoder of the CLIP model to emphasize visual relationship contexts and object categories guided by natural language expressions. CLIP introduces a novel approach to conveying visual concepts. Text embeddings of the corresponding image are obtained using an intuitively simple prompt. For each class, we employ the template ``a photo of \texttt{[CLS]},'' where \texttt{[CLS]} represents the class label. Additionally, to represent text excluding the region of the target object in the image, we generate a background text embedding using the prompt ``a photo of \texttt{background}''. Through these prompts, we generate text embeddings $T=\left\{t^{fg}, {t^{bg}} \right\} $ using the CLIP text encoder, where $t^{fg}\in \mathbb{R}^{C \times d_t}$ and $t^{bg} \in \mathbb{R}^{1 \times d_t}$ denote foreground and background text embeddings, respectively. Further, $d_t$ denotes channel dimension, and $C$ represents the number of classes. 

\subsection{Dense Alignment between Image and Text Representations}
In two modalities—the corresponding visual features and text embeddings—we expect semantic coherence at various levels of granularity. Following the aforementioned process, we obtain representations associated with the foreground and background of each modality. For visual features related to objects, the similarity to the foreground text embedding should be high, whereas the similarity with the background text embedding should be low. This principle also applies conversely, especially regarding background visual features. Building on this interaction, we employ a dual-level alignment strategy. To align the representations, we project the visual features from the visual encoder and the text embeddings from the text encoder into a unified vision-language embedding space. \\
{\bf Global Implicit Alignment.} We provide comprehensive textual guidance by aligning visual features with text embeddings. Global Implicit Alignment (GIA) is achieved through cross-contrastive learning between the class token of the projected visual features and the projected text embeddings. The cosine similarity $\mathtt{sim}(\cdot)$ is computed between the class token $v_{cls} \in \mathbb{R}^{1 \times d}$ and the foreground and background text embeddings, denoted as $t^{fg} \in \mathbb{R}^{C \times d}$ and $t^{bg} \in \mathbb{R}^{1 \times d}$, respectively, where $d$ denotes the projected feature dimension. As the class token $v_{cls}$ aggregates global semantics, the objective is to maximize the similarity to $t^{fg}$ and minimize the similarity to $t^{bg}$. The objective of GIA is represented as follows:
\begin{equation}
    \mathcal{L}_{im} = -\sum_{c=1}^{C} y_{c}\cdot \log(\mathtt{sim}(v_{cls} , t^{fg}_c)) -\log (1-\mathtt{sim}(v_{cls} , t^{bg})),
\label{eq:eq5}
\end{equation}
where $y_c$ denotes the image-level label for $c$-th class. Although class tokens have different semantics, GIA enables class tokens to effectively aggregate related visual global representations by incorporating background text embeddings. However, depending on GIA, it may neglect localization details, as it might not fully capture the entire object region due to limited interaction with the class token.\\
{\bf Local Explicit Alignment.} To capture the entire object region and enhance localization details, we introduce Local Explicit Alignment (LEA). In contrast to GIA using the class token that aggregates semantic attributes, we leverage patch tokens that effectively preserve local information. For cross-contrastive learning, we explore various levels of granularity by utilizing foreground and background features from both images and texts.

We project flattened patch tokens, $v^{fg}_p\in\mathbb{R}^{N^2\times d}$ and $v^{bg}_p\in\mathbb{R}^{N^2\times d}$, obtained by applying the object-aware mask $M_{obj}$. We consider $t^{fg}$ as the positive pair and $t^{bg}$ as the negative pair for $v^{fg}_p$. Conversely, for $v^{bg}_p$, we match $t^{bg}$ as the positive pair and $t^{fg}$ as the negative pair. 

\begin{equation}
    s^{fg}_{+} = \sum_{c=1}^C y_c\cdot \mathtt{sim}(v_{p}^{fg}, t^{fg}_c),\quad  s^{fg}_{-} = \mathtt{sim}(v_{p}^{fg}, t^{bg}),
\end{equation}
\begin{equation}
    s^{bg}_{+} = \mathtt{sim}(v_{p}^{bg},t^{bg}), \quad s^{bg}_{-} = \sum_{c=1}^C y_c\cdot \mathtt{sim}(v_{p}^{bg}, t^{fg}_c),
\end{equation}
where $y_c$ denotes the image-level label for $c$-th class.
Similar to other studies~\cite{ru2023token, du2022weakly, xie2022c2am}, we adopt the InfoNCE~\cite{oord2018representation} loss as the objective.
\begin{flalign}
\mathcal{L}_{ex} = -\frac{1}{N^2} \sum_{i=1}^{N^2} \log \frac{\footnotesize\text{exp}({s^{fg}_{+_i}/\tau})}{\footnotesize\text{exp}({s^{fg}_{+_i}/\tau}) +\footnotesize\text{exp}({s^{fg}_{-_i}/\tau})} \nonumber\\  
-\lambda \frac{1}{N^2} \sum_{i=1}^{N^2} \log \frac{\footnotesize\text{exp}({s^{bg}_{+_i}/\tau})}{\footnotesize\text{exp}({s^{bg}_{+_i}/\tau})+\footnotesize\text{exp}({s^{bg}_{-_i}/\tau})}, 
\label{eq:eq8}
\end{flalign}
where $\lambda$ is the hyperparameter for balancing, and $\tau$ is the temperature factor. We empirically set them to 0.001 and 1.0, respectively. Explicitly cross-aligning visual features and text embeddings across foreground and background representations enables the identification of specific regions based on the target class embedding. Additionally, by separating patch tokens based on the foreground and background and aligning them with the text embeddings, the classification network can more precisely distinguish objects and backgrounds in the image.

\subsection{Network Training}
We apply the global max pooling operation to the visual features obtained from the classifier of the classification network and the intermediate CAMs to compute the logits. Subsequently, we compute the multi-label soft margin loss, represented as $\mathcal{L}_{cls}$ and $\mathcal{L}_{inter}$, for each logit. The overall loss in the proposed method includes $\mathcal{L}_{im}$, $\mathcal{L}_{ex}$, $\mathcal{L}_{cls}$, and $\mathcal{L}_{inter}$. To address the problem of over-smoothing in the ViT, we incorporate the patch tokens contrast loss $\mathcal{L}_{ptc}$ following ~\cite{ru2023token}. The loss for training the model is defined as follows:
\begin{equation}
    \mathcal{L} = \mathcal{L}_{cls} + \mathcal{L}_{inter} + \lambda_{i} \mathcal{L}_{im} + \lambda_{e} \mathcal{L}_{ex} + \lambda_p \mathcal{L}_{ptc},
\label{eq:total_loss}
\end{equation}
where $\lambda_i$, $\lambda_e$, and $\lambda_p$ are hyperparameters. Furthermore, DALNet is designed as a single-stage WSSS framework. Additionally, to improve boundary alignment, we integrate the pixel-adaptive refinement (PAR) module~\cite{ru2022learning}. This module refines the pseudo labels generated by DALNet, and these refined pseudo labels then supervise the segmentation head. For semantic segmentation, we employ the widely used cross-entropy loss, denoted by $\mathcal{L}_{seg}$. Similar to previous single-stage WSSS studies~\cite{pan2022learning,ru2022weakly,ru2023token}, we integrate an additional regularization loss~\cite{tang2018regularized} $\mathcal{L}_{reg}$ to ensure the spatial consistency of the predicted segmentation masks. The total loss for semantic segmentation is represented as follows:
\begin{equation}
    \mathcal{L}_{total} = \mathcal{L} + \mathcal{L}_{seg} + \mathcal{L}_{reg}.
\end{equation}


\section{Experiments}
\subsection{Dataset and Evaluation Metric}
We evaluate the proposed method on the PASCAL VOC 2012~\cite{everingham2010pascal} segmentation benchmark, consisting of 20 foreground classes and one background class. Following common practice in semantic segmentation, we take additional annotations from SBD~\cite{hariharan2011semantic}. The official dataset split contains 10,582 images for training, 1,449 for validation, and 1,456 for testing. We also employ the MS COCO 2014~\cite{lin2014microsoft} dataset, with 81 classes, 82k training, and 40k validation images. During the training stage, we only use image-level labels. To evaluate segmentation results, we employ the mean intersection over union (mIoU) metric.

\begin{figure}[t!]

  \centering
   \includegraphics[width=0.7\linewidth]{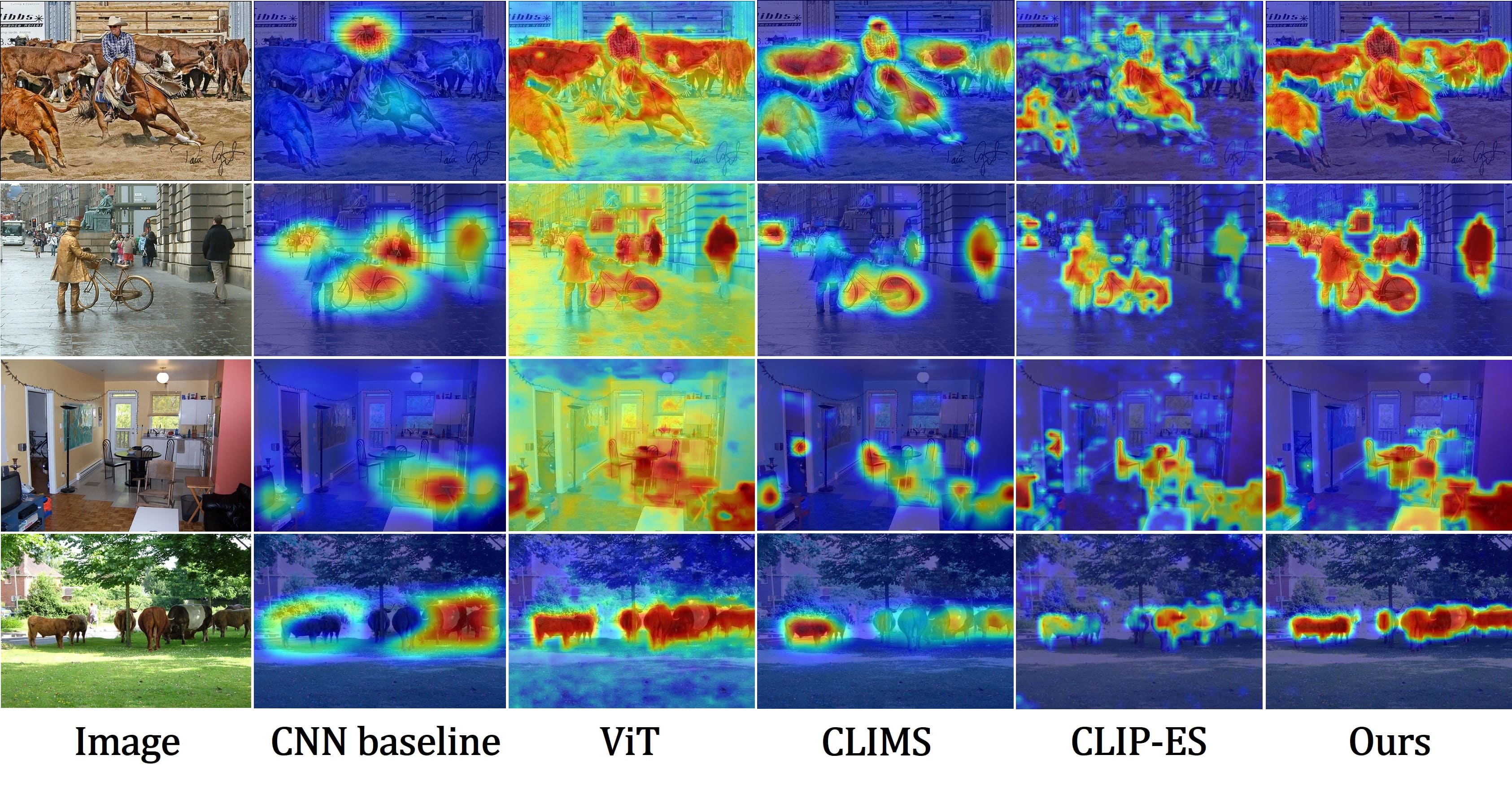}
   \caption{Visualization results of CAMs. We generate CAMs using CNN baseline, ViT, CLIMS~\cite{xie2022clims}, CLIP-ES~\cite{lin2023clip} and our proposed DALNet.}
   \label{fig:fig3}
\end{figure}

\begin{table}[t!]
  \caption{Performance comparison of the initial CAMs on the PASCAL VOC dataset~\cite{everingham2010pascal}. $\dagger$ denotes pre-trained parameters from ImageNet-21k~\cite{ridnik2021imagenet}.}
  \centering
  \begin{tabular}{lclccc}
    \toprule
\multicolumn{1}{l|}{{\bf Method}} & \multicolumn{1}{c|}{{\bf Backbone}}      & \multicolumn{1}{c|}{{\bf Train}} & {\bf Val} \\ \hline
\multicolumn{1}{l|}{RRM~\cite{zhang2020reliability}\textcolor[gray]{0.5}{\tiny{AAAI'20}}}       & \multicolumn{1}{c|}{WR38}& \multicolumn{1}{c|}{-}& {65.4}\\
\multicolumn{1}{l|}{1Stage~\cite{araslanov2020single}\textcolor[gray]{0.5}{\tiny{CVPR'20}}}    & \multicolumn{1}{c|}{WR38}& \multicolumn{1}{c|}{66.9}& {65.3}\\           
\multicolumn{1}{l|}{SLRNet~\cite{pan2022learning}\textcolor[gray]{0.5}{\tiny{IJCV'22}}}    & \multicolumn{1}{c|}{WR38}& \multicolumn{1}{c|}{67.1}& {66.2}\\
\multicolumn{1}{l|}{AFA~\cite{ru2022learning}\textcolor[gray]{0.5}{\tiny{CVPR'22}}}       & \multicolumn{1}{c|}{MiT-B1}& \multicolumn{1}{c|}{68.7}& {66.5}\\   
\multicolumn{1}{l|}{ViT-PCM~\cite{rossetti2022max}\textcolor[gray]{0.5}{\tiny{ECCV'22}}}& \multicolumn{1}{c|}{ViT-B$^{\dagger}$}& \multicolumn{1}{c|}{67.7}& {66.0}\\
\multicolumn{1}{l|}{CLIMS~\cite{xie2022clims}\textcolor[gray]{0.5}{\tiny{CVPR'22}}}      & \multicolumn{1}{c|}{R50}& \multicolumn{1}{c|}{56.6}&{-}\\ 
\multicolumn{1}{l|}{MTCformer~\cite{xu2022multi}\textcolor[gray]{0.5}{\tiny{CVPR'22}}}      & \multicolumn{1}{c|}{ViT-B}& \multicolumn{1}{c|}{61.7}&{-}\\ 
\multicolumn{1}{l|}{AMN~\cite{lee2022threshold}\textcolor[gray]{0.5}{\tiny{CVPR'22}}}      & \multicolumn{1}{c|}{R50}& \multicolumn{1}{c|}{62.1}&{-}\\ 
\multicolumn{1}{l|}{MMCST~\cite{xu2023learning}\textcolor[gray]{0.5}{\tiny{CVPR'23}}}      & \multicolumn{1}{c|}{ViT-B$^{\dagger}$}& \multicolumn{1}{c|}{66.3}&{-}\\ 
\multicolumn{1}{l|}{CLIP-ES~\cite{lin2023clip}\textcolor[gray]{0.5}{\tiny{CVPR'23}}}      & \multicolumn{1}{c|}{ViT-B$^{\dagger}$}& \multicolumn{1}{c|}{70.8}&{-}\\ 
\multicolumn{1}{l|}{ToCo~\cite{ru2023token}\textcolor[gray]{0.5}{\tiny{CVPR'23}}}      & \multicolumn{1}{c|}{ViT-B$^{\dagger}$}& \multicolumn{1}{c|}{73.6}&{72.3}\\ 
\multicolumn{1}{l|}{Ours}                                                 & \multicolumn{1}{c|}{ViT-B}& \multicolumn{1}{c|}{71.9}& {70.2}\\              
\multicolumn{1}{l|}{Ours$^{\dagger}$}                                     & \multicolumn{1}{c|}{ViT-B$^{\dagger}$}& \multicolumn{1}{c|}{{\bf 75.2}}&{{\bf 73.1}}\\   
    \bottomrule
  \end{tabular}%
  \label{tab:tab1}
\end{table}

\subsection{Experimental Settings}
{\bf Network Architectures.} We use ViT-base (ViT-B)~\cite{dosovitskiy2020image} as the visual encoder initiated from pre-trained weights obtained from ImageNet~\cite{ridnik2021imagenet}. We employ bilinear interpolation to adapt the positional embedding, ensuring compatibility with the input size and enabling the handling of different-sized images. We utilize the frozen text encoder from the CLIP model. The segmentation head is constructed with a 1$\times$1 prediction layer and two 3$\times$3 convolutional layers. \\
{\bf Implementation Details.} We use an AdamW optimizer. The learning rate gradually increases to 6e-5 for the first 1,500 iterations and then decays with a polynomial scheduler. The warm-up rate is set to 1e-6, and the decay rate is 0.9. In the experiments conducted on the PASCAL VOC dataset, we set the batch size to 4 and the total number of iterations to 20k. On the MS COCO dataset, the network is trained for 80k iterations with a batch size of 8. To generate object-aware masks, we utilize the features of transformer's 10-th block in ~\cref{sec:features}. The background threshold $\beta$ is set at 0.5. We use (1.0, 1.0, 0.2) for the loss weight factors ($\lambda_i$, $\lambda_e$, $\lambda_p$) in ~\cref{eq:total_loss}. Additionally, we employ the data augmentation and multi-crop approaches described in~\cite{caron2021emerging}. Following common practices in semantic segmentation~\cite{chen2017deeplab}, we apply dense conditional random fields (CRF) processing and multi-scale testing during the inference stage.
\begin{table}[t!]
  \caption{Comparison of the semantic segmentation results on the PASCAL VOC validation and testing sets. Sup. indicates the type of supervision ($\mathcal{I}$: image-level label; $\mathcal{S}$: saliency map; $\mathcal{L}$: image-level language). Net. denotes the backbone network (for single-stage methods) and the semantic segmentation network (for multi-stage methods). $\dagger$ denotes the use of ImageNet-21k~\cite{ridnik2021imagenet} pre-trained parameters.}
  \setlength{\tabcolsep}{4pt}
  \centering
  \begin{tabular}{lllll}
    \toprule
\multicolumn{1}{l|}{} & \multicolumn{1}{c|}{{\bf Sup.}} & \multicolumn{1}{c|}{{\bf Net.}} & \multicolumn{1}{c|}{{\bf Val}} & {\bf Test} \\ \hline
\textit{{\bf Multi-stage WSSS methods}} &                         &                        &                          & \\   
\multicolumn{1}{l|}{RIB~\cite{hariharan2011semantic}\textcolor[gray]{0.5}{\scriptsize{NeurIPS'21}}}   &\multicolumn{1}{c|}{$\mathcal{I+S}$}& \multicolumn{1}{c|}{DL-V2}& \multicolumn{1}{c|}{70.2} & {70.0}\\
\multicolumn{1}{l|}{EPS~\cite{lee2021railroad}\textcolor[gray]{0.5}{\scriptsize{CVPR'21}}}      &\multicolumn{1}{c|}{$\mathcal{I+S}$}& \multicolumn{1}{c|}{DL-V2}& \multicolumn{1}{c|}{71.0} & {71.8}\\
\multicolumn{1}{l|}{L2G~\cite{jiang2022l2g}\textcolor[gray]{0.5}{\scriptsize{CVPR'22}}}      &\multicolumn{1}{c|}{$\mathcal{I+S}$}& \multicolumn{1}{c|}{DL-V2}& \multicolumn{1}{c|}{72.1} & {71.7}\\
\multicolumn{1}{l|}{RCA~\cite{zhou2022regional}\textcolor[gray]{0.5}{\scriptsize{CVPR'22}}}      &\multicolumn{1}{c|}{$\mathcal{I+S}$}& \multicolumn{1}{c|}{DL-V2}& \multicolumn{1}{c|}{72.2} & {72.8}\\
\multicolumn{1}{l|}{Du \etal~\cite{du2022weakly}\textcolor[gray]{0.5}{\scriptsize{CVPR'22}}} &\multicolumn{1}{c|}{$\mathcal{I+S}$}& \multicolumn{1}{c|}{DL-V2}& \multicolumn{1}{c|}{72.6} & {73.6}\\
\multicolumn{1}{l|}{ReCAM~\cite{chen2022class}\textcolor[gray]{0.5}{\scriptsize{CVPR'22}}}    &\multicolumn{1}{c|}{$\mathcal{I}$}  & \multicolumn{1}{c|}{DL-V2}& \multicolumn{1}{c|}{68.4} & {68.2}\\
\multicolumn{1}{l|}{VWL~\cite{ru2022weakly}\textcolor[gray]{0.5}{\scriptsize{IJCV'22}}}      &\multicolumn{1}{c|}{$\mathcal{I}$}  & \multicolumn{1}{c|}{DL-V2}& \multicolumn{1}{c|}{69.2} & {69.2}\\
\multicolumn{1}{l|}{W-OoD~\cite{lee2022weakly}\textcolor[gray]{0.5}{\scriptsize{CVPR'22}}}    &\multicolumn{1}{c|}{$\mathcal{I}$}  & \multicolumn{1}{c|}{WR38} & \multicolumn{1}{c|}{70.7} & {70.1}\\
\multicolumn{1}{l|}{MCTformer~\cite{xu2022multi}\textcolor[gray]{0.5}{\scriptsize{CVPR'22}}}&\multicolumn{1}{c|}{$\mathcal{I}$}  & \multicolumn{1}{c|}{WR38} & \multicolumn{1}{c|}{71.9} & {71.6}\\
\multicolumn{1}{l|}{ESOL~\cite{li2022expansion}\textcolor[gray]{0.5}{\scriptsize{NeurIPS'22}}}  &\multicolumn{1}{c|}{$\mathcal{I}$}  & \multicolumn{1}{c|}{DL-V2}& \multicolumn{1}{c|}{69.9} & {69.3}\\
\multicolumn{1}{l|}{LPCAM~\cite{chen2023extracting}\textcolor[gray]{0.5}{\scriptsize{CVPR'23}}}    &\multicolumn{1}{c|}{$\mathcal{I}$}  & \multicolumn{1}{c|}{DL-V2}& \multicolumn{1}{c|}{70.1} & {70.4}\\
\multicolumn{1}{l|}{FPR~\cite{chen2023fpr}\textcolor[gray]{0.5}{\scriptsize{ICCV'23}}}      &\multicolumn{1}{c|}{$\mathcal{I}$}  & \multicolumn{1}{c|}{DL-V2}& \multicolumn{1}{c|}{70.3} & {70.1}\\
\multicolumn{1}{l|}{CLIMS~\cite{xie2022clims}\textcolor[gray]{0.5}{\scriptsize{CVPR'22}}}    &\multicolumn{1}{c|}{$\mathcal{I+L}$}& \multicolumn{1}{c|}{DL-V2}& \multicolumn{1}{c|}{69.3} & {68.2}\\
\multicolumn{1}{l|}{CLIP-ES~\cite{lin2023clip}\textcolor[gray]{0.5}{\scriptsize{CVPR'23}}}  &\multicolumn{1}{c|}{$\mathcal{I+L}$}& \multicolumn{1}{c|}{DL-V2}& \multicolumn{1}{c|}{71.1} & {71.4}\\
\multicolumn{1}{l|}{MMCST~\cite{xu2023learning}\textcolor[gray]{0.5}{\scriptsize{CVPR'23}}}    &\multicolumn{1}{c|}{$\mathcal{I+L}$} & \multicolumn{1}{c|}{WR38}& \multicolumn{1}{c|}{72.2} & {72.2}\\ \hline
\textit{{\bf Single-stage WSSS methods}} &                        &                           &                          & \\  
\multicolumn{1}{l|}{RRM~\cite{zhang2020reliability}\textcolor[gray]{0.5}{\scriptsize{AAAI'20}}}      &\multicolumn{1}{c|}{$\mathcal{I}$}& \multicolumn{1}{c|}{WR38}& \multicolumn{1}{c|}{62.6}& {62.9}\\
\multicolumn{1}{l|}{1Stage~\cite{araslanov2020single}\textcolor[gray]{0.5}{\scriptsize{CVPR'20}}}   &\multicolumn{1}{c|}{$\mathcal{I}$}   & \multicolumn{1}{c|}{WR38}& \multicolumn{1}{c|}{62.7}& {64.3}\\
\multicolumn{1}{l|}{AFA~\cite{ru2022learning}\textcolor[gray]{0.5}{\scriptsize{CVPR'22}}}      &\multicolumn{1}{c|}{$\mathcal{I}$}   & \multicolumn{1}{c|}{MiT-B1}& \multicolumn{1}{c|}{66.0}& {66.3}\\              
\multicolumn{1}{l|}{SLRNet~\cite{pan2022learning}\textcolor[gray]{0.5}{\scriptsize{IJCV'22}}}   &\multicolumn{1}{c|}{$\mathcal{I}$}   & \multicolumn{1}{c|}{WR38}& \multicolumn{1}{c|}{67.2}& {67.6}\\
\multicolumn{1}{l|}{ViT-PCM~\cite{rossetti2022max}\textcolor[gray]{0.5}{\scriptsize{ECCV'22}}}  &\multicolumn{1}{c|}{$\mathcal{I}$}  & \multicolumn{1}{c|}{ViT-B}& \multicolumn{1}{c|}{70.3} & {70.9}\\
\multicolumn{1}{l|}{ToCo~\cite{ru2023token}\textcolor[gray]{0.5}{\scriptsize{CVPR'23}}}     &\multicolumn{1}{c|}{$\mathcal{I}$}   & \multicolumn{1}{c|}{ViT-B$^{\dagger}$}& \multicolumn{1}{c|}{71.1}&{72.2}\\              
\multicolumn{1}{l|}{Ours}                                                &\multicolumn{1}{c|}{$\mathcal{I+L}$} & \multicolumn{1}{c|}{ViT-B}& \multicolumn{1}{c|}{{\bf 71.4}}& {{\bf 71.4}}\\               
\multicolumn{1}{l|}{Ours$^{\dagger}$}                                    &\multicolumn{1}{c|}{$\mathcal{I+L}$} & \multicolumn{1}{c|}{ViT-B$^{\dagger}$}& \multicolumn{1}{c|}{{\bf 74.5}}&{{\bf 74.9}}\\   
    \bottomrule
  \end{tabular}%
  \label{tab:tab2}
\end{table}

\subsection{Experimental Results}
\begin{figure*}[t!]
 \centering
 \includegraphics[width=1.0\linewidth]{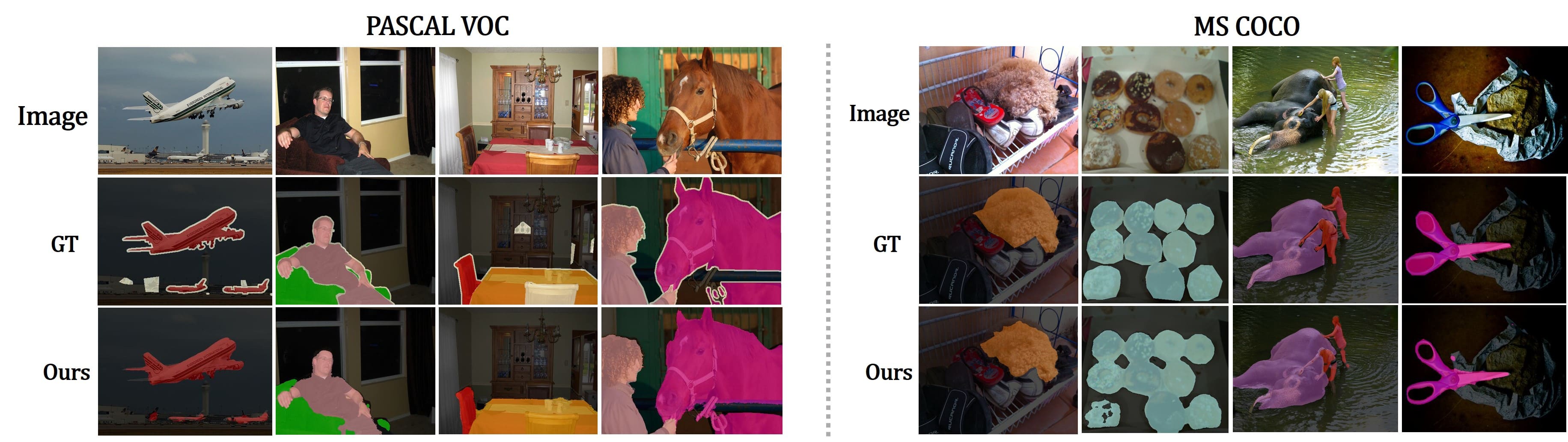}
  \caption{Visualization results of semantic segmentation on the PASCAL VOC and MS COCO datasets.}
  \label{fig:fig4}
\end{figure*}

{\bf Quality of Pseudo Labels.} We visualize CAMs using the proposed method, which activates multiple objects within the image, as shown in \cref{fig:fig3}. Our approach offers more precise object localization across the entire object region compared to recent methods like CLIMS~\cite{xie2022clims} and CLIP-ES~\cite{lin2023clip}, which rely on text guidance. We conduct a quantitative evaluation of the training and validation sets of the PASCAL VOC dataset. We compare our DALNet with existing WSSS methods in \cref{tab:tab1}. To ensure a fair comparison, we include results using pre-trained ImageNet-1k weights (DeiT~\cite{touvron2021training}), considering the initial pre-training of ViT on ImageNet-21k. Our findings demonstrate that our method generates CAMs with mIoU comparable to or exceeding those of recent methods.\\
{\bf Semantic Segmentation Results.} The semantic segmentation outcomes are detailed in \cref{tab:tab2} and \cref{tab:tab3} from experiments conducted on the PASCAL VOC and MS COCO datasets. Notably, our proposed method significantly outperforms on the PASCAL VOC dataset, achieving mIoU of 74.5\% and 74.9\% on the validation and testing sets, respectively. Furthermore, we achieve a mIoU of 44.4\% on the MS COCO validation dataset, demonstrating effective performance compared to multi-stage methods and superior results compared to single-stage methods. We emphasize that our study is dedicated to advancing single-stage WSSS with a unified objective. As shown in \cref{fig:fig4}, the visualized semantic segmentation masks demonstrate the activation of various objects, providing a comprehensive understanding of the images. This suggests that incorporating textual information as guidance enhances object localization.
\begin{table}[t!]
  \caption{Comparison of the semantic segmentation results on the MS COCO~\cite{lin2014microsoft} validation set. Net. denotes the backbone network (for single-stage methods) and the semantic segmentation network (for multi-stage methods). $\dagger$ indicates the use of ImageNet-21k~\cite{ridnik2021imagenet} pre-trained parameters.}
  \centering
  \begin{tabular}{lll||lll}
    \toprule
& \multicolumn{1}{|c|}{{\bf Net.}}      & {{\bf Val}}& & \multicolumn{1}{|c|}{{\bf Net.}}      & {{\bf Val}}               \\ \hline

\multicolumn{3}{l||}{\textit{{\bf \normalsize{Multi-stage WSSS methods}}}} & \multicolumn{3}{l}{\textit{{\bf Single-stage WSSS methods}}} \\ 

\multicolumn{1}{l|}{SEAM~\cite{wang2020self}\textcolor[gray]{0.5}{\tiny{CVPR'20}}}     & \multicolumn{1}{c|}{DL-V1}& {31.9} &
\multicolumn{1}{l|}{AFA~\cite{ru2022learning}\textcolor[gray]{0.5}{\tiny{CVPR'22}}}       & \multicolumn{1}{c|}{MiT-B1}& {38.9}\\           

\multicolumn{1}{l|}{RIB~\cite{hariharan2011semantic}\textcolor[gray]{0.5}{\tiny{NeurIPS'21}}}   & \multicolumn{1}{c|}{DL-V2}& {43.8} &
\multicolumn{1}{l|}{SLRNet~\cite{pan2022learning}\textcolor[gray]{0.5}{\tiny{ECCV'22}}}    & \multicolumn{1}{c|}{WR38}& {35.0}\\           

\multicolumn{1}{l|}{RCA~\cite{zhou2022regional}\textcolor[gray]{0.5}{\tiny{CVPR'22}}}      & \multicolumn{1}{c|}{DL-V2}& {36.8} & 
\multicolumn{1}{l|}{ToCo~\cite{ru2023token}\textcolor[gray]{0.5}{\tiny{CVPR'23}}}      & \multicolumn{1}{c|}{ViT-B$^{\dagger}$}& {42.3}\\

\multicolumn{1}{l|}{SIPE~\cite{chen2022self}\textcolor[gray]{0.5}{\tiny{CVPR'22}}}     & \multicolumn{1}{c|}{DL-V2}& {40.6} &
\multicolumn{1}{l|}{Ours}                                     & \multicolumn{1}{c|}{ViT-B}&{{\bf 42.7}}\\

\multicolumn{1}{l|}{MCTformer~\cite{xu2022multi}\textcolor[gray]{0.5}{\tiny{CVPR'22}}}& \multicolumn{1}{c|}{WR38}& {42.0} &
\multicolumn{1}{l|}{Ours$^{\dagger}$}                                     & \multicolumn{1}{c|}{ViT-B$^{\dagger}$}&{{\bf 44.4}}\\
    \bottomrule
  \end{tabular}
  \label{tab:tab3}
\end{table}
\subsection{Ablation and Analysis}
\begin{figure}[t]
 \centering
  \includegraphics[height=4cm]{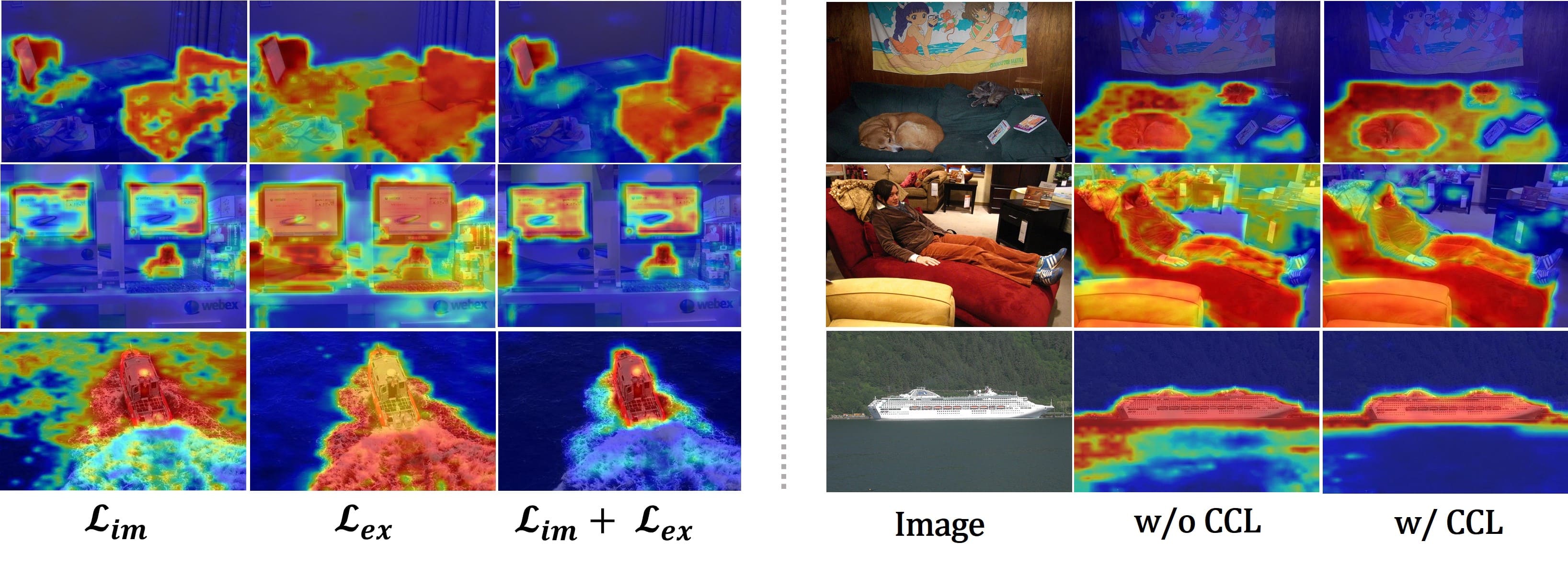}
  \caption{\textbf{(Left)} Visualization of initial CAMs with different loss function configurations. \textbf{(Right)} Visualization results for the CAMs on the PASCAL VOC dataset without and with cross-contrastive learning (CCL).}
  \label{fig:fig5}
\end{figure}

{\bf Configurations of Loss Function.} In our ablation study, we explore the impact of the proposed method. The results are detailed in \cref{tab:tab4}, including the results of initial CAMs on the PASCAL VOC validation set. For dense alignment between images and texts, we employ $\mathcal{L}_{im}$ (GIA) and $\mathcal{L}_{ex}$ (LEA). Without cross-contrastive learning, the mIoU of CAMs is 64.1\% when only $\mathcal{L}_{im}$ is applied. Applying only $\mathcal{L}_{ex}$ results in CAMs mIoU of 65.1\%. Combining $\mathcal{L}_{im}$ and $\mathcal{L}_{ex}$ significantly improves the mIoU of CAMs to 66.9\%, indicating the best performance. As shown in \cref{fig:fig5} (Left), relying solely on global ($\mathcal{L}_{im}$) or local ($\mathcal{L}_{ex}$) information either fails to activate the integral object regions or results in over-activation. Our dense alignment approach ($\mathcal{L}_{im} + \mathcal{L}_{ex}$) effectively enhances dense localization attained by integrating global and local information. \\
\begin{table}[t]
       \caption{Comparison of initial CAMs (mIoU) with different loss function configurations and cross-contrastive learning (CCL) on the PASCAL VOC validation set.} 
        \setlength{\tabcolsep}{6pt}
        \centering
               \begin{tabular}{c|cc|cc}
\toprule
\textbf{Method} & {\bf $\mathcal{L}_{im}$\scriptsize{(GIA)}}\quad & {\bf $\mathcal{L}_{ex}$\scriptsize{(LEA)}} &  {\bf w/o CCL} & {\bf w/ CCL} \\ \hline
\multirow{3}{*}{ViT}  &\checkmark &         & {64.1} & {67.9}\\
                               &           & \checkmark & {65.1}&{68.3}  \\
                               &\checkmark & \checkmark & {66.9}&{70.2} \\ 
    \bottomrule
    \end{tabular}
    \label{tab:tab4}
\end{table}
{\bf Analysis of Cross-contrastive Learning.} We facilitate cross-contrastive learning between visual features and text embeddings. In \cref{tab:tab4}, we compare the performance of initial CAMs using cross-contrastive learning (CCL). CCL pairs visual foreground with target class embeddings as positives and background text embeddings as negatives. Simultaneously, it pairs the visual background with background text embeddings as positives and target class embeddings as negatives. Using only the foreground as positives (w/o CCL), the mIoU of the CAMs is 66.9\%. By incorporating the foreground and background (w/ CCL), the mIoU of the CAMs increases to 70.2\%. These results demonstrate the effectiveness of our cross-contrastive learning approach. It indicates that incorporating background visual features and text embeddings can enhance performance by addressing missing regions and mitigating distractions, ultimately improving focus on the target object. By utilizing specific classes and the background prompts, we align the background text embedding with areas outside the foreground in images. The ``\texttt{background}'' term is broadly defined to encompass diverse elements beyond image objects. Notably, it can produce promising outcomes by eliminating the need for specific background categorization and additional prior efforts. Additionally, \cref{fig:fig5} (Right) presents a visualization of CAMs results, demonstrating the effectiveness of our proposed method in suppressing over-activated background areas and compensating for the under-activation in object areas within the CAMs.\\
\begin{figure}[t]
  \centering
   \includegraphics[height=4cm]{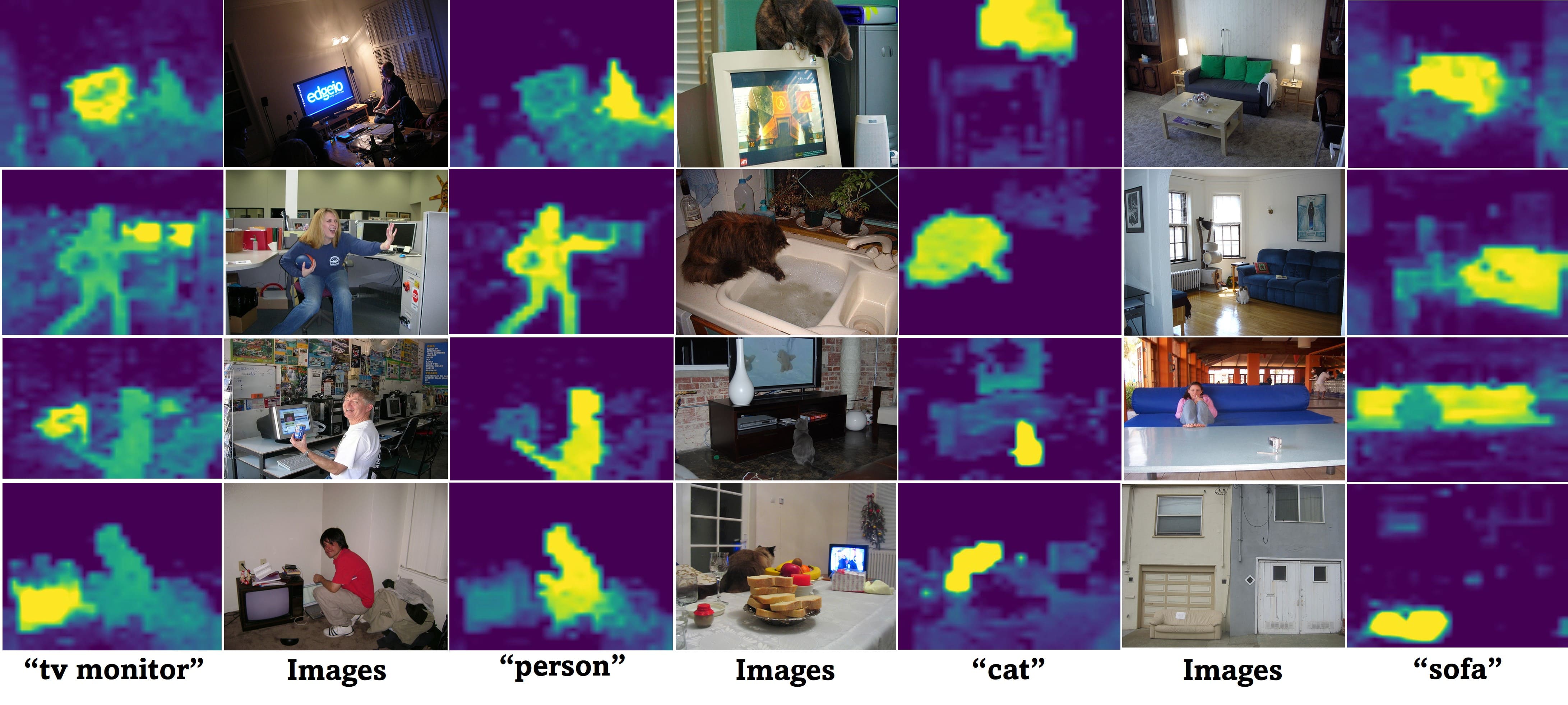}
   \caption{Visualization of similarity coherence between image and text representations for each class.}
   \label{fig:fig6}
\end{figure}
\textbf{Similarity Coherence.} To validate the effectiveness of DALNet, which combines GIA and LEA, we present similarity coherence between image patch tokens and categorical text embeddings. Visualization results using the PASCAL VOC training set are shown in \cref{fig:fig6}. Our proposed method enables the identification of specific regions based on the target class embedding. Specifically, local contrastive learning with object-ware mask can enhance the similarity of patch tokens corresponding to the target class text embedding. Our focus on WSSS leverages the categories defined within the dataset. However, there are approaches in VLP that use open-vocabulary categories to learn the similarity coherence between text and image~\cite{mukhoti2023open, xu2022groupvit, wang2022cris}. Unlike previous studies that depend on additional annotation masks or fine-tuning modules, learning processes based solely on text embeddings and image features provide insights for improving generalization in our future exploration.

\section{Conclusion}
In this paper, we address the challenges of discriminative CAMs and WSSS that rely on image-level labels. We present a novel approach, Dense Alignment Learning Network (DALNet), which leverages text guidance at different levels of granularity. DALNet consists of two strategies: Global Implicit Alignment (GIA) for semantic information and Local Explicit Alignment (LEA) for spatial details. Based on this dual-level alignment strategy, DALNet aligns foregrounds across modalities and distinguishes them from their backgrounds through cross-contrastive learning. Extensive experiments show that DALNet improves dense localization by combining global and local information. Furthermore, we demonstrate that by utilizing straightforward prompts with specific class labels and the ``\texttt{background}'' term, DALNet enhances focus on target objects by simultaneously activating missing regions and suppressing distractions. Our experimental findings validate its effectiveness, achieving outstanding performance on the PASCAL VOC and MS COCO datasets, even as a single-stage WSSS framework. 
\section*{Acknowledgements}
This research was supported by the Basic Science Research Program through the National Research Foundation of Korea (NRF) funded by the Ministry of Education (NRF-2022R1C1C1008534, Contribution Rate: 25\%); the Institute for Information \& Communications Technology Planning \& Evaluation (IITP) grant funded by the Korea government (MSIT) (No. 2021-0-01341, Artificial Intelligence Graduate School Program, Chung-Ang University, Contribution Rate: 25\%); and the Culture, Sports and Tourism R\&D Program through the Korea Creative Content Agency grant funded by the Ministry of Culture, Sports and Tourism in 2024 (Project Name: Developing Professionals for R\&D in Contents Production Based on Generative AI and Cloud, Project Number: RS-2024-00352578, Contribution Rate: 50\%).

%
%
\bibliographystyle{splncs04}

\end{document}